\documentclass[10pt,twocolumn,letterpaper]{article}

\usepackage{cvpr}      % To produce the REVIEW version
\usepackage{graphicx}
\usepackage{amsmath}
\usepackage{amssymb}
\usepackage{booktabs}
\usepackage{algorithm}
\usepackage{multirow}
\usepackage{multicol}
\usepackage{url}
\usepackage[pagebackref,breaklinks,colorlinks]{hyperref}

\usepackage[capitalize]{cleveref}
\crefname{section}{Sec.}{Secs.}
\Crefname{section}{Section}{Sections}
\Crefname{table}{Table}{Tables}
\crefname{table}{Tab.}{Tabs.}

\def\cvprPaperID{4} % *** Enter the CVPR Paper ID here
\def\confName{CVPR}
\def\confYear{2023}

\begin{document}

\title{Prompting Large Language Models to Reformulate Queries for Moment Localization}

\author{Wenfeng Yan$^1$\quad
Shaoxiang Chen$^2$ \quad
Zuxuan Wu$^1$\quad
Yu-Gang Jiang$^1$\\
$^1$Shanghai Key Lab of Intelligent Information Processing, \\School of Computer Science, Fudan University\\
$^2$Meituan\\
{\tt\small wfyan22@m.fudan.edu.cn, \{sxchen13, zxwu, ygj\}@fudan.edu.cn}}

\maketitle

%%%%%%%%% ABSTRACT
\begin{abstract}
%In this paper, we present a query reformulation based solution for the Ego4D Natural Language Queries (NLQ) Challenge in CVPR 2023. 
The task of moment localization is to localize a temporal moment in an untrimmed video for a given natural language query. Since untrimmed video contains highly redundant contents, the quality of the query is crucial for accurately localizing moments, i.e., the query should provide precise information about the target moment so that the localization model can understand what to look for in the videos.
However, the natural language queries in current datasets may not be easy to understand for existing models. For example, the Ego4D dataset uses question sentences as the query to describe relatively complex moments. While being natural and straightforward for humans, understanding such question sentences are challenging for mainstream moment localization models like 2D-TAN. 
Inspired by the recent success of large language models, especially their ability of understanding and generating complex natural language contents, in this extended abstract, we make early attempts at reformulating the moment queries into a set of instructions using large language models and making them more friendly to the localization models. 
%Given a video clip and a natural language query, the goal of this challenge is to localize a temporal moment where the answer to the query can be deduced. 
%Consider that the given natural language query may bring inadequate information compared to the informative video clip, we propose to use large language models (LLMs) to reformulate the query into a series of localization instructions.
\end{abstract}

%%%%%%%%% BODY TEXT
\section{Introduction}
\label{sec:intro}

% 逻辑：
% 1. 任务介绍，这个任务很难
% 2. 我们可以通过分解步骤来实现更好的效果
% 3. 人工分解非常耗时，最近LLM的文本处理能力很强（介绍一下chatgpt，以及其他开源模型，LLAMA，Vicuna），且可以通过prompt实现特定功能的处理
% 4. 引出我们要用LLM来重写query实现更好的效果

Given a video clip and a natural language query, the goal of the Ego4D NLQ task \cite{grauman2022ego4d} is to localize a temporal moment of the video clip where the answer to the query can be deduced. In contrast to the traditional NLQ benchmark datasets such as ActivityNet Captions \cite{krishna2017dense} and TACoS \cite{rohrbach2014coherent}, Ego4D NLQ uses a question sentence (\eg, \emph{``Did I turn off the cooker after I fried the meat?"}) rather than a narrative sentence, which requires stronger reasoning ability and deeper comprehension of spatio-temporal relationship between objects, places, and people in the videos. Moreover, compared to the duration of full video clips, most temporal windows are extremely short, further making the task a challenging “needle in the haystack” search problem.

Since the visual content of long video clips is highly redundant, the natural language query may not contain sufficient information for precise temporal localization. To address this problem, we propose to reformulate the query into a series of localization instructions so that the model can learn to perform temporal localization step-by-step and reduce its burden of performing complex reasoning. For instance, consider the query ``\emph{Did I turn off the cooker after I fried the meat?}", it is natural to first localize the moment where the person fried the meat, then localize the moment after that with the cooker, where the person interacts with the cooker and the answer can be found. Thus the query can be reformulated into a better format such as ``\emph{find the moment where I fried the meat, next find the moment after this where I interacted with the cooker and may turned off the cooker.}" However, such reformulation procedure is time-consuming and expensive if performed by humans. 
\begin{figure}[t]
  \centering
   \includegraphics[width=1.0\linewidth]{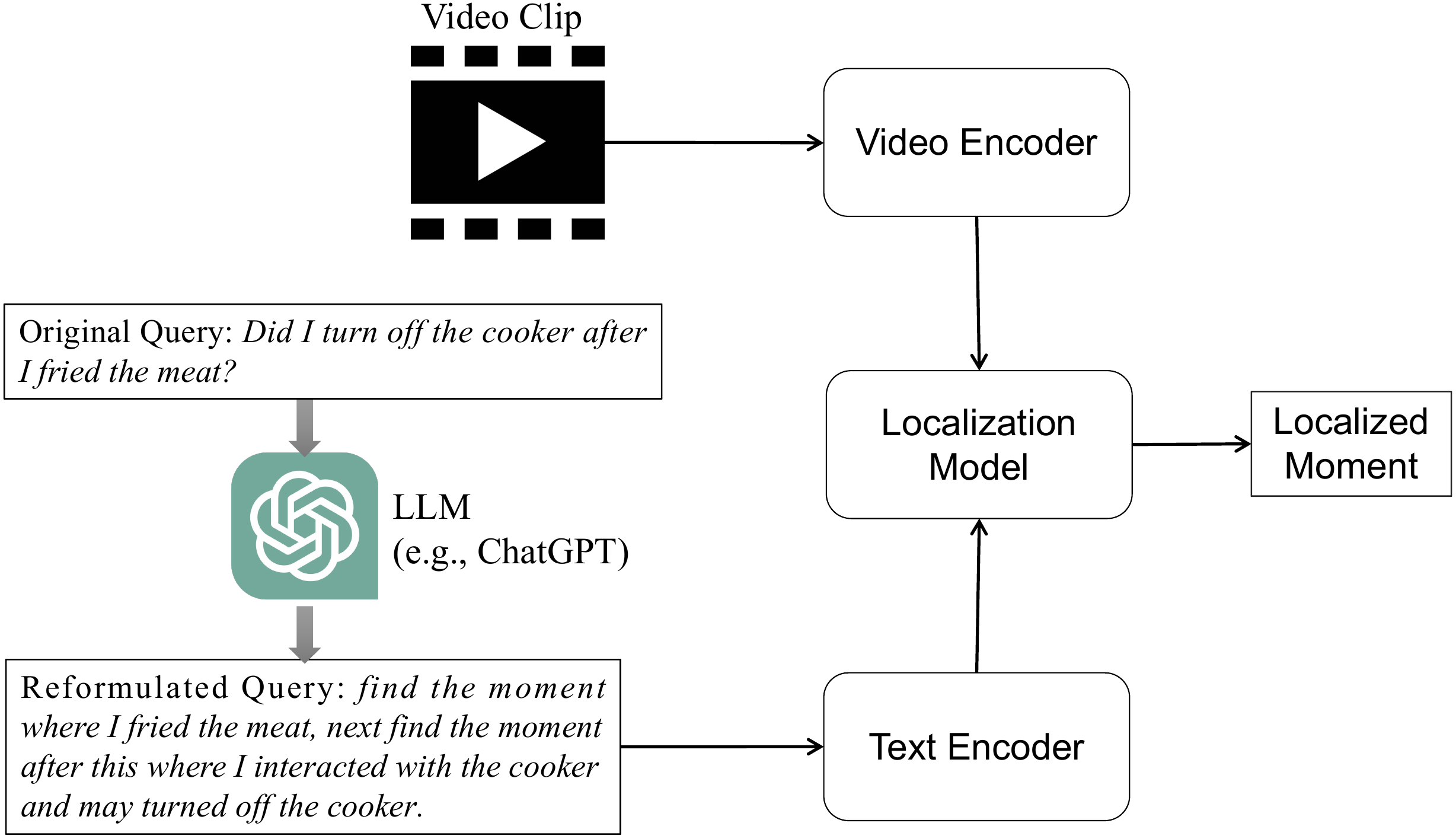}

   \caption{The overall framework of our approach.}
   \label{fig:frame}
\end{figure}

Recently, LLMs have shown great potential in natural language processing, i.e., understanding complex user queries and generating responses that align with human preference. One of the most significant recent breakthroughs is ChatGPT~\cite{chatgpt}, which is built upon OpenAI's GPT~\cite{radford2018improving,radford2019language,brown2020language,2303.08774} family models and finetuned to be able to interact with users in a conversational manner and exhibits step-by-step reasoning abilities. 
More importantly, ChatGPT can be prompted to act by user-defined rules and give corresponding responses in a conversation. Some following open-source LLMs with fewer parameters also have close performance with ChatGPT, such as LLaMA~\cite{touvron2023llama}, Vicuna~\cite{vicuna2023}. 
Inspired by LLMs' abilities, we attempt to prompt LLMs to reformulate the queries (especially Ego4D's question sentences) into a more friendly format for the localization model. Since this is a work in progress, we present early results and analysis.

\section{Methodology}
In this section, we first introduce the basic formulation of the Ego4D NLQ task. Then, we introduce how we prompt LLMs to reformulate the natural language queries.
\subsection{Task Formulation}
Given a video clip $V$ and a natural language query $Q$, the Ego4D NLQ task aims to localize the best-matching temporal moment $M$ where the answer to query can be deduced. Specifically, we denote the video clip with $T$ consecutive frames as $V=\{f_t\}_{t=1}^{T}$ and the sentence query with $L$ words as $Q=\{s_i\}_{i=1}^{L}$. Each video clip is associated with multiple temporal sentence annotations $A=\{(Q_j,\tau_j^s,\tau_j^e)\}_{j=1}^{N}$, where $N$ is the number of sentence annotations. The task is to predict the best-matching temporal moment $(\tau_j^s,\tau_j^e)$ for each sentence.

\subsection{Prompting LLMs for Better Queries}

%Large language models that are pre-trained on massive amounts of text data and finetuned to align with human preference have demonstrated promising performance in a wide range of tasks, including language translation, text summarization, and question answering. Particularly, ChatGPT, which was trained on a very large web corpus, has strong ability in text understanding and logical reasoning, leading to an increasing number of following applications and research.
We attempt to automatically reformulate the natural queries using LLMs, by prompting them with a carefully-engineered prompt. The prompt requires the LLM to act as an assistant that takes the original query as input, comprehends it, and reformulates it into a series of instructions for the localization model. 
We also provide the query templates of the original queries in the prompt to help the LLM understand the queries.
As shown in Figure~\ref{fig:frame}, to obtain better results, we use the powerful ChatGPT to reformulate the query into a series of localization instructions (by utilizing the OpenAI API, smaller and open-source LLMs will also be tested in the future).
The reformulated queries will be the new input to an existing localization model (such as 2D-TAN).
We post our prompt in Algorithm \ref{alg:prompt}.
Note that the reformulated query may be significantly detailed while not changing the original meaning of the query, thus a redesign of the localization model is needed in the future.
% 形式化描述大模型
% prompt 贴到 algorithm https://openaccess.thecvf.com/content_CVPR_2020/papers/He_Momentum_Contrast_for_Unsupervised_Visual_Representation_Learning_CVPR_2020_paper.pdf

\begin{algorithm}[!htb]\scriptsize
\caption{Prompt for LLMs.}
\texttt{You are Eva, a super intelligent assistant that help users locate moments in videos via natural language queries.\\\\You are:\\- helpful and friendly\\- not able to directly access the video's content\\- decompose a complex event query into a series of logically coherent actions\\- good at understanding user's intent and extract the core steps from the query in order to answer the user's question\\\\You can use an external tool named Locator, which is able to locate moments in videos given detailed natural language queries.\\\\The user will ask a question about objects, places, and people in an ego-centric video, and the key to answer the question is to first locate relevant moments given the query.\\\\Your goal is to reformulate the query into a series of instructions for the Locator.\\\\There are some templates for user's query as following:\\- Where is object X before / after event Y?\\- Where is object X?\\- What did I put in X?\\- How many X's? (quantity question)\\- What X did I Y?\\- In what location did I see object X ?\\- What X is Y?\\- State of an object\\- Where is my object X?\\- Where did I put X?\\- Who did I interact with when I did activity X?\\- Who did I talk to in location X?\\- When did I interact with person with role X?\\\\Here are some examples:\\Example 1:\\query: What did I sprinkle on the cooking pan?\\output: find the moment when I sprinkled something on the cooking pan.\\Example 2:\\query: Did I turn off the cooker after I fried the meat?\\output: find the moment when I fried the meat, next find the moment after this with the cooker (I may turn off the cooker).\\
\\
Now reformulate this query \{USER\_INPUT\}:\\}
\label{alg:prompt}
\end{algorithm}
\section{Experiments}
To verify the effectiveness of the query reformulation, we conduct localization experiments based on 2D-TAN \cite{zhang2020learning} with the original queries and the reformulated queries for comparison.
\subsection{Experimental Setting}

\textbf{Dataset.}
Ego4D is a recently collected massive-scale egocentric video dataset and benchmark suite, containing 3,670 hours of daily life activity video spanning hundreds of scenarios (household, outdoor, workplace, leisure, etc.). 
The Ego4D NLQ task is an official subtask that aims to localize the temporal moment of a video clip given a natural language query. The average duration of video clips is about 8.25 minutes. In total, there are over 13.8K queries and 4.5K queries in the training and validation set, respectively. 

\textbf{Model Details.}
We apply 2D-TAN with a sliding window method (a baseline provided in the original paper\cite{grauman2022ego4d}) on the Ego4D NLQ dataset. The core idea of 2D-TAN is to construct a two-dimensional temporal map consisting of moment candidates and retrieve the target moment on the resulting score map. Since the video clips are very long compared to the short temporal window, we further break down the video clip into a number of overlapping windows before inputting it into the 2D-TAN model.

For the video clip, we use Ego4D’s provided features extracted by the SlowFast \cite{feichtenhofer2019slowfast} network. For the language query, we use the BERT \cite{devlin2018bert} model to obtain the sentence embedding. For the sliding window method, we choose 40s and 20s as the window duration and the window stride respectively. During training, we only use windows that intersect with the ground-truth moment, but we use all windows for testing. We train the model for 20 epochs with a learning rate of 0.0001 and report the performance on the validation set.

\subsection{Results of Query Reformulation}
\begin{figure}[t]
  \centering
   \includegraphics[width=1.0\linewidth]{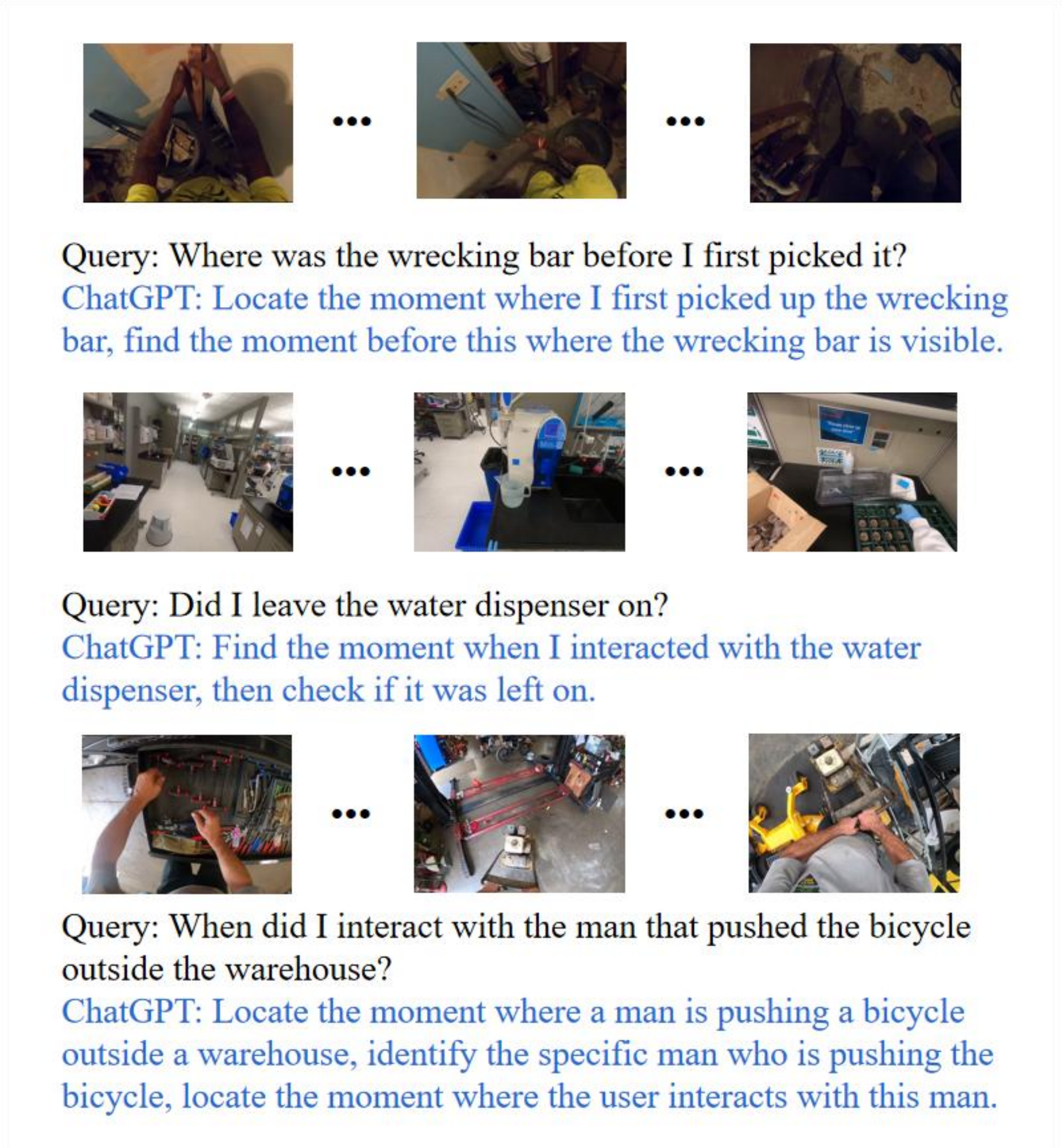}

   \caption{Some examples of the reformulated queries.}
   \label{fig:query}
\end{figure}
Figure~\ref{fig:query} shows some examples of the reformulated queries by ChatGPT. 
We can see that ChatGPT can decompose the query into a series of detailed localization instructions without changing the original meaning, which may lead to more precise moment localization compared to the original question sentence and is easier for the localization model to comprehend. 
On average, there are approximately 7.56 words and 14.91 words in the original queries and the reformulated queries, respectively.

\subsection{Results of Moment Localization}
\begin{table}[!htb]
  \centering
  \begin{tabular}{ccccccc}
    \toprule
    \multicolumn{3}{c}{\multirow{2}{*}{Method}}& \multicolumn{2}{c}{IoU=0.3 (\%)}& \multicolumn{2}{c}{IoU=0.5 (\%)}\\
    \multicolumn{3}{c}{}&R@1&R@5&R@1&R@5\\  
    \midrule 
    \multicolumn{3}{c}{2D-TAN}& 4.57 & 12.88 & 2.86 & 8.11\\
    \multicolumn{3}{c}{2D-TAN-R} & 4.26 & 12.90 & 2.51 & 7.67\\
    \bottomrule
  \end{tabular}
  \caption{Performance comparison on the validation set.}
  \label{tab:result}
\end{table}
We preliminarily evaluated the effect of query reformulation on Ego4D NLQ validation set. Specifically, we adopt the standard metric “R@$n$, IoU=$m$” for evaluation, which is defined as the percentage of query sentences that have at least one grounding prediction whose IoU with ground-truth is larger than $m$ among top-$n$ results. 
Table \ref{tab:result} summarizes the results, where 2D-TAN refers to the experimental result with the original queries and 2D-TAN-R refers to the experimental result with the reformulated queries. 
From the preliminary results, we can observe that the baseline performance of 2D-TAN is quite low, and query reformulation leads to a slight performance drop for the 2D-TAN model. The reason may be that 2D-TAN encodes the entire sentence into a single feature vector, which cancels the benefit of the multiple localization instructions of the reformulated queries. And with longer sentences, the encoded single feature vector is harder to preserve key information needed to execute step-by-step localization.
Furthermore, 2D-TAN is inherently not designed for step-by-step localization and its language encoding module is too simple to handle such set of localization instructions.
In future work, we will attempt to make better use of the reformulated queries for precise moment localization by designing a novel and more suitable model.

\section{Conclusion}
In this extended abstract, we make an attempt at using the recently-developed powerful large language model to reformulate natural language queries into a step-by-step localization instructions for the moment localization problem. 
Specifically, we perform preliminary experiments using the ChatGPT to reformulate the sentence queries of the Ego4D NLQ task and test the performances of the popular 2D-TAN model.
Our initial findings indicates that current models like 2D-TAN are not suitable for the reformulated queries and a redesign is needed in a future work.

%%%%%%%%% REFERENCES
{\small
\bibliographystyle{unsrt}
\bibliography{egbib}
}

\end{document}